\documentclass{llncs}

\usepackage{amsmath,amssymb,amsfonts}
\usepackage{graphicx}
\usepackage[draft]{fixme}
\usepackage[table]{xcolor}

\usepackage[pagebackref]{hyperref} 
\hypersetup{colorlinks=true} 
\usepackage[boxruled]{algorithm2e} 
\usepackage{paralist}                           
\usepackage{bm}
\usepackage{wrapfig}
\usepackage{float}
\usepackage{multirow}
\usepackage{booktabs}
\usepackage{nicefrac}
\usepackage{nth}
\usepackage[bf,small,skip=0pt]{caption}
\usepackage{subcaption}
\usepackage{wrapfig}
\usepackage{dsfont}
\usepackage{tikz}
\usetikzlibrary{shapes,arrows,backgrounds,positioning}

\DeclareMathOperator{\id}{id}

\usepackage{mathtools}
\makeatletter
\newcommand{\mytag}[2]{%
  \text{#1}%
  \@bsphack
  \protected@write\@auxout{}%
         {\string\newlabel{#2}{{#1}{\thepage}}}%
  \@esphack
}
\makeatother

\usepackage{caption}
\graphicspath{{figures/}}

\title{Fast Predictive Image Registration}
\author{Xiao Yang$^1$, Roland Kwitt$^3$, Marc Niethammer$^{1,2}$}
\institute{$^1$Department of Computer Science, UNC Chapel Hill, USA\\
$^2$Biomedical Research Imaging Center, UNC Chapel Hill, USA\\
$^3$Department of Computer Science, University of Salzburg, Austria}
\begin{document}
\maketitle
\begin{abstract}
  We present a method to predict image deformations based on patch-wise image appearance. Specifically, we design a patch-based deep encoder-decoder network which learns the pixel/voxel-wise mapping between image appearance and registration parameters. Our approach can predict general deformation parameterizations, however, we focus on the large deformation diffeomorphic metric mapping (LDDMM) registration model. By predicting the LDDMM momentum-parameterization we retain the desirable theoretical properties of LDDMM, while reducing computation time by orders of magnitude:  combined with patch pruning, we achieve a $1500\mathsf{x}$/$66\mathsf{x}$ speed up compared to GPU-based optimization for 2D/3D image registration. Our approach has better prediction accuracy than predicting deformation or velocity fields and results in diffeomorphic transformations. Additionally, we create a Bayesian probabilistic version of our network, which allows evaluation of deformation field uncertainty through Monte Carlo sampling using dropout at test time. We show that deformation uncertainty highlights areas of ambiguous deformations. We test our method on the OASIS brain image dataset in 2D and 3D. 
\end{abstract}
\section{Introduction}
Image registration is a critical medical image analysis task to provide spatial correspondences. A prominent application is atlas-to-image registration, commonly used for atlas-based segmentations or population analyses. Image registration is typically cast as an optimization problem, which can be especially computationally demanding for non-parametric diffusive, elastic, or fluid models~\cite{modersitzki2004} such as LDDMM~\cite{beg2005}. Recently, approaches to {\it predict} registration parameters have been proposed: resulting deformations can be (i) used directly or (ii) to initialize an optimizer. However, high parameter dimensionality and the non-linearity between image appearance and the registration parameters makes predictions challenging. Chou et al.~\cite{chou20132d} propose a multi-scale linear regressor, which is restricted to the prediction of affine transformations and low-rank approximations of non-rigid deformations. For complex deformable registrations, Wang et al.~\cite{Wang201561} use image-template key point matching with sparse learning and a subsequent interpolation to a dense deformation field via radial basis functions. While effective, this method is dependent on proper key point selection. In \cite{Tian2015}, a semi-coupled dictionary learning technique is used to jointly model image appearance and the deformation parameters; however, assuming a linear relationship between image appearance and deformation parameters only, which is overly restrictive.\\
\indent Optical flow~\cite{deepflow,flownet} and affine transforms~\cite{Miao2016} have been computed via deep learning. Here, we explore a deep learning regression model\footnote{Other regression models could of course be used as well.} for parameter prediction for non-parametric image-registration from image patches. 

\vskip0.5ex
\noindent
\textbf{Contribution.} {\it Convenient parameterization:} Using the momentum parameterization for LDDMM shooting~\cite{Vialard:2012}, we retain the mathematical properties of LDDMM under patch-based prediction, e.g., we guarantee diffeomorphic transforms.
{\it Fast computation:} Using a sliding window with a large stride and patch pruning to predict the momentum, we achieve dramatic speed-ups compared to a direct optimization approach while maintaining high prediction accuracy. {\it 3. Atlas-based formulation:} In contrast to generic 
optical flow approaches, we use an atlas-based viewpoint. This allows us to predict the momentum in a fixed atlas coordinate system and hence within a consistent tangent space. {\it Uncertainty quantification:} We provide a Bayesian model from which estimates for parameter uncertainty and consequentially deformation map uncertainty can be obtained. This information can be used, e.g., for uncertainty-based smoothing~\cite{simpson2011longitudinal}, surgical treatment planning, or for direct uncertainty visualizations.

\vskip0.5ex
\noindent\textbf{Organization.}
Sec.~\ref{sec:LDDMM} reviews the registration parameterization of LDDMM. Sec.~\ref{sec:network} introduces our network structure and our strategy for speeding up deformation prediction. Sec.~\ref{sec:experiment} presents experimental results for both 2D and 3D brain images from the OASIS~\cite{OASIS} brain image data set, and discusses the generality of our approach, as well as possible improvements and extensions.

\vspace{-0.2cm}
\section{Initial Momentum LDDMM Parameterization} 
\label{sec:LDDMM}
Given a source image $S$ and a target image $T$, a {\it time-dependent} deformation map $\Phi:\mathbb{R}^d \times \mathbb{R} \rightarrow \mathbb{R}^d$, maps between the coordinates of $S$ and $T$, at time $t=1$, i.e., $S(\Phi(x,1)) = T(x)$; $d$ denotes the spatial dimension. In the LDDMM shooting formulation~\cite{Vialard:2012}, the initial momentum vector field $m_0$ is the registration parameter from which $\Phi$ can be computed. In fact, by integrating the geodesic equations~\eqref{eqn:forward}, the complete spatio-temporal transformation, $\Phi(x,t)$ is determined. 
The initial momentum is the dual of the initial velocity $v_0$, which is an element in the reproducing kernel Hilbert space $V$, and they are connected by a positive-definite, self-adjoint smoothing operator $K$ as $v = Km$ and $m = Lv$, where $K=L^{-1}$. The energy for the shooting formulation of LDDMM is
\begin{equation}
    E(m_0) = \langle m_0, Km_0\rangle + \frac{1}{\sigma^2}||S\circ \Phi(1) - T||^2,\quad \text{s.t.}
    \label{eqn:momentum_energy}
\end{equation}
\begin{equation}
\begin{split}
m_t + \text{ad}_v^*m = 0,\, m(0)=m_0, \, \Phi_t + D \Phi v = 0,\,  \Phi(0)=\id, \, m - Lv = 0,
\label{eqn:forward}
\end{split}
\end{equation}
where $\id$ is the identity map, $\text{ad}^*$ is the dual of the negative Jacobi-Lie bracket of vector fields: $\text{ad}_vw = Dvw - Dwv$, $D$ denotes the Jacobian, and $\sigma>0$. \emph{Our goal} is to predict the initial momentum $m_0$ given the source and target images patch by patch. We will show, in Sec.~\ref{sec:experiment}, that this is a convenient parameterization as (i) the momentum does not need to be smooth, but is compactly supported at image edges and (ii) the velocity is obtained by smoothing the momentum via $K$. Hence, smoothness does not need to be  considered in the prediction step, but is imposed {\it after} prediction. $K$ governs the theoretical properties of LDDMM: in particular, a strong enough $K$ assures diffeomorphic transformations, $\Phi$~\cite{beg2005}. Hence, by predicting $m_0$, we retain the theoretical properties of LDDMM. Furthermore, patch-wise prediction of alternative parameterizations of LDDMM, such as the initial velocity, or directly predicting displacements, is difficult in homogeneous image regions, as these regions provide no information to guide the registration. As the 
\emph{momentum} in such regions is zero\footnote{For image-based LDDMM the momentum is $m=\lambda \nabla I$, where $\lambda$ is a scalar-valued momentum field and $I$ is the image. Hence, $m=0$ in uniform areas of $I$.}, no such issues arise.

\vspace{-0.2cm}
\section{Network Structure} 
\label{sec:network}

\begin{figure}[t!]
\begin{center}
\includegraphics[width=1.00\textwidth]{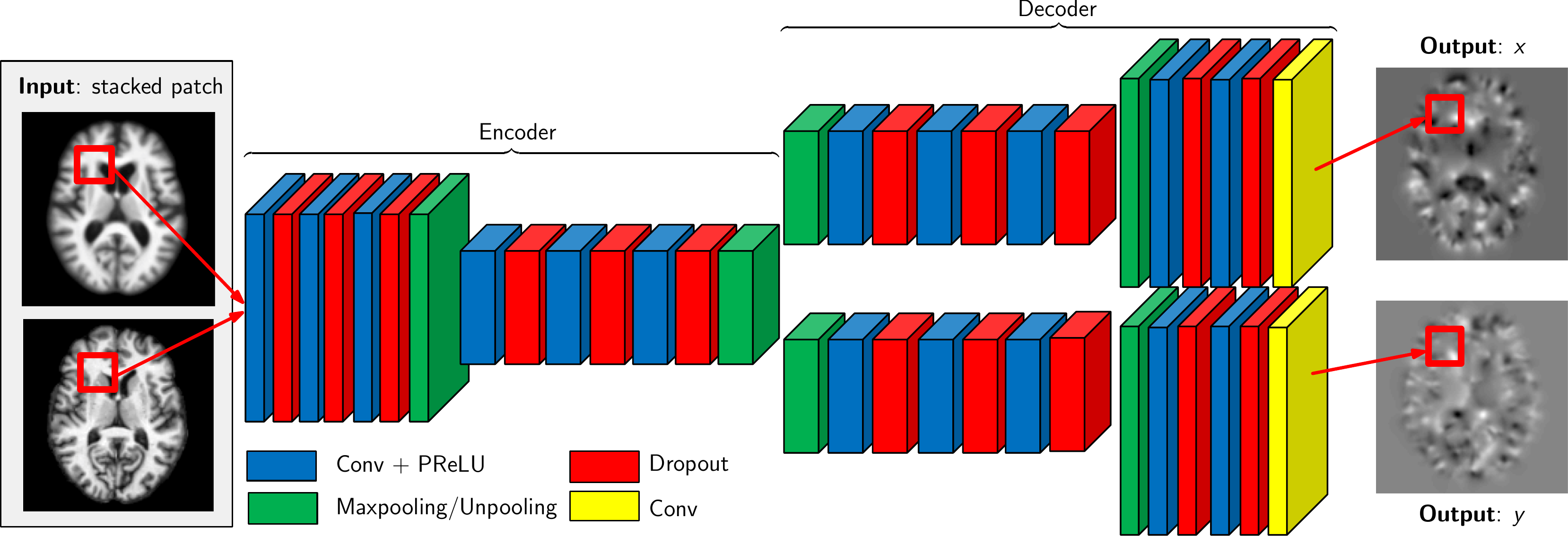}
\vskip2ex
\caption{Bayesian probabilistic network structure (for 2D images): The inputs are 2-layer stacked patches from the moving image and fixed image at the same location. The output is the initial momentum prediction of the patches in $x$ and $y$ spatial directions. For a deterministic version of the network, we simply remove all dropout layers. For a 3D image network we increase the number of decoders to 3 and use volumetric layers.\label{fig:network}}
\end{center}
\vspace{-0.2cm}
\end{figure}

Fig. \ref{fig:network} shows the structure of our initial momentum prediction network. We first discuss our deterministic version \textit{without} dropout layers, then introduce the probabilistic network using dropout. We focus the discussion on 2D images for notational simplicity, but also implement and experiment with 3D networks by using volumetric layers and adding an additional decoder for the \nth{3} dimension.

\vskip0.5ex
\noindent
\textbf{Deterministic Network.} In the 2D version of the network, the input is a two layer $15\times15$ image patch, where the two layers come from the fixed and the moving image, resp., taken at the same location, and the network output is the initial momentum prediction patch for $x$ and $y$ directions. Our network consists of two parts: the \emph{encoder} and the \emph{decoder}. In the {\it encoder}, we create a VGG-style~\cite{Simonyan14c} network with 2 blocks of three $3\times3$ convolutional layers with PReLU~\cite{PReLU} activations, and $2\times2$ maxpooling layers with a step size of 2 at the end of each two blocks. The number of features in the convolutional layers is 128 for the large image scale block, and 256 for the smaller one. The {\it decoder} contains two parallel decoders sharing input generated from the encoder; each decoder's structure is the inverse of the encoder, except for using max-unpooling layers with the pooling layers' indices, and no non-linearity layer at the end. Unpooling layers help retain image boundary detail, which is important for initial momentum prediction. During training, we use the $L_1$ difference for network output evaluation. To compute a momentum prediction for the \emph{whole} image, we use a sliding window and patch averaging in the 
overlapping areas.
We use two (three) independent decoders to predict the initial momentum in 2D (3D) as, experimentally, such a network structure is much easier to train than a network with one large decoder 
to predict the initial momentum in all dimensions simultaneously. In our experiments, such a combined network easily got stuck in poor local minima. Our independent decoder network can be regarded as a multi-task network,  where each decoder predicts initial momentum for a single dimension.


\vskip0.5ex
\noindent\textbf{Bayesian Probabilistic Network Using Dropout.} We extend our network to a Bayesian probabilistic network by using \emph{dropout}~\cite{srivastava14a}. This can be regarded as approximate variational inference for a Bayesian network with Bernoulli distributions over the network's weights~\cite{Gal2015Bayesian}. 
Given a 2-layer image patch $X$ and the corresponding initial momentum patch $Y$, we determine the weights $W$ of the convolutional layers so that given input $X$, our network is likely to generate the target output $Y$. We define the likelihood $p(Y|W, X)$  of the network output via the $L_1$ difference. Our goal is to find the posterior of the weights $W$, i.e., $p(W |Y, X)\propto (p(Y|W, X) p(W)) /p(Y)$, where $p(W)\sim \mathcal{N}(0, I)$ is the prior of $W$, and $p(Y)$ is constant. Since this posterior is generally unknown, we use a variational posterior $q(W)$ to approximate the true posterior by minimizing the Kullback-Leibler (KL) divergence $D_{\text{KL}}(q(W)~||~p(W|X, Y)))$. When using dropout for convolutional layers, the variational posterior $q(W_i)$ for the $i$th convolutional layer with $K_i\times K_i$ weight matrix can be written as~\cite{Gal2015Bayesian}
\begin{equation}
q(W_i) = W_i\cdot \text{diag}([z_{i, j}]_{j = 1}^{K_i}),\quad z_{i, j} \sim \text{Bernoulli}(p_i),
\end{equation}
where $z_{i, j}$ is a Bernoulli random variable modeling dropout with probability $p_i$, randomly setting the $j$th node in the $i$th layer to 0. The variational parameter is the network weight $W_i$. The variational posterior for all network weights $W$ is then $q(W) = \prod_{i}q(W_i)$. According to~\cite{Gal2015Bayesian}, we minimize KL-divergence by adding dropout layers after all convolutional layers except for the final output, as shown in Fig.~\ref{fig:network}, and train the network using stochastic gradient descent. During test time, we keep the dropout layers, and evaluate the posterior by Monte Carlo sampling of the network given fixed input data. We use the sample mean as our final initial momentum, from which we compute the deformation by integrating Eqs.~\eqref{eqn:forward} forward. We calculate the \textsl{deformation} variance by integrating Eqs.~\eqref{eqn:forward} for each initial momentum sample separately. We set $p_i=0.3$.

\vskip0.5ex
\noindent
\textbf{Speeding Up Whole Image Deformation Prediction.} As we predict the whole-image initial momentum patch-by-patch, computation speed is proportional to the number of patches. 
We use two techniques to reduce the number of patches/image, thereby increasing computation speed: First, we perform \emph{patch pruning} by ignoring all patches from the background of both the moving and the target image; this can be done, since the initial momentum for the constant background should be zero. Second, we use a large pixel/voxel stride (e.g., 14 for $15\times15$ patches) for the sliding window. This is reasonable, because of the compact support (around edges) of the initial momentum and the shift-invariance property of pooling/unpooling layers. These two techniques reduce the number of predicted patches/image by $99.5\%$ for the $128\times128$ 2D images and by $99.97\%$ for the $128\times128\times128$ 3D images, at a negligible loss of accuracy (cf. Sec. \ref{sec:experiment}).
\begin{table}[t!]
\scriptsize
\centering
\begin{subtable}{1\linewidth}
\begin{tabular}{|r|c|c|c|c|c|c|c|c|}
\hline
& \multicolumn{7}{c|}{\textbf{2D Test Case Deformation Error} [pixel]} & \bm{$\text{\textbf{det}} J > 0$}\\ \hline
\textsl{Data Percentile} & 0.3\% & 5\% & 25\% & 50\% & 75\% & 95\% & 99.7\% &\\ \hline
Affine & 0.0925 & 0.3779 & 0.9207 & 1.4741 & 2.1717 & 3.4606 & 5.4585 & N/A\\ \hline
SCDL, 1000 dictionary & 0.0819 & 0.337 & 0.8156 & 1.3078 & 1.9368 & 3.1285 & 4.7948 & 100\%\\ \hline
D, velocity, stride 1 & 0.0228  & 0.0959 & 0.2453 & 0.4343 & 0.7354 & 1.4664 & 2.9768 & 100\%\\ \hline
D, velocity, stride 14+PR & 0.027 & 0.1123 & 0.2878 & 0.5075 & 0.8605 & 1.75 & 3.6172 & 76\%\\ \hline
D, displacement, stride 1 & 0.0215 & 0.0897 & 0.2332 & 0.416 & 0.7064 & 1.429 & 2.9462 & 90\%\\ \hline
D, displacement, stride 14+PR & 0.0252 & 0.107 & 0.2786 & 0.4955 & 0.8047 & 1.7298 & 3.7327 & 0\% \\ \hline\hline
\textbf{D, stride 1} & 0.0194 & 0.0817 & 0.2035 & 0.3436 & 0.5618 & 1.1395 & 2.473 & 100\%\\ \hline
\textbf{D, stride 14+PR} & 0.0221 & 0.0906 & 0.2244 & 0.375 & 0.6057 & 1.2076 & 2.6731 & 100\%\\ \hline
\textbf{P, stride 1, 50 samples} & {\cellcolor{green!30}{\bf 0.0185}} & {\cellcolor{green!30}{\bf 0.0787}} & {\cellcolor{green!30}{\bf 0.1953}} & 
{\cellcolor{green!30}{\bf 0.3261}} & {\cellcolor{green!30}{\bf 0.5255}} & {\cellcolor{green!30}{\bf 1.0745}} & {\cellcolor{green!30}{\bf 2.3525}} & 100\%  \\ \hline
\textbf{P, stride 14+PR, 50 samples} & 0.0209 & 0.0855 & 0.2123  & 0.351  & 0.5556
 & 1.1133 &  2.5678 &  100\% \\ \hline
\end{tabular}
\vskip2ex
\end{subtable}
\begin{subtable}{1\linewidth}
\begin{tabular}{|r|c|c|c|c|c|c|c|c|}
\hline
& \multicolumn{7}{c|}{\textbf{3D Test Case Deformation Error} [voxel]} & \bm{$\text{\textbf{det}} J > 0$} \\ \hline
\textsl{Data Percentile} & 0.3\% & 5\% & 25\% & 50\% & 75\% & 95\% & 99.7\%& \\ \hline
Affine & 0.0821 & 0.2529 & 0.5541 & 0.8666 & 1.2879 & 2.1339 & 3.7032 & N/A\\ \hline
\textbf{D, stride 7} & {\cellcolor{green!30}{\bf 0.0128}} & {\cellcolor{green!30}{\bf 0.0348}} & {\cellcolor{green!30}{\bf 0.0705}} & {\cellcolor{green!30}{\bf 0.1072}} & {\cellcolor{green!30}{\bf 0.1578}} & {\cellcolor{green!30}{\bf 0.2663}} & {\cellcolor{green!30}{\bf 0.5049}} & 100\% \\ \hline
\textbf{D, stride 14+PR} & 0.0146 & 0.0403 & 0.0831 & 0.1287 & 0.194 & 0.351 & 0.6896 & 100\% \\ \hline
\textbf{P, stride 14+PR, 50 samples} & 0.0151 & 0.0422 & 0.0876 & 0.1363 & 0.2051 & 0.3664 & 0.8287 & 100\% \\ \hline
\end{tabular}
\end{subtable}
\vskip4ex
\caption{\label{table:2D3D} Test results for 2D (\emph{top}) and 3D (\emph{bottom}). \emph{SCDL}: semi-coupled dictionary learning; \emph{D}: deterministic network; 
\emph{P}: probabilistic network; \emph{stride}: stride length for sliding window for whole image prediction; \emph{velocity}: predicting initial velocity; \emph{displacement}: predicting displacement field; \emph{PR}: patch pruning. The column \bm{$\text{\textbf{det}} J > 0$} shows the ratio of test cases with positive definite Jacobian determinant of the deformation map. Our initial momentum networks (and the best results) are highlighted in \textbf{bold}.}
\vspace{-0.2cm}
\end{table}

\vspace{-0.2cm}
\section{Experiments}
\label{sec:experiment}
\vspace{-0.2cm}
We evaluate our method using 2D ($128\times128$) and 3D ($128\times128\times128$) images of the OASIS longitudinal dataset~\cite{OASIS}. We use the first scan of all subjects, resulting in 150 images. The 2D slices are extracted from the same axial slice of the 3D images after affine registration. We randomly picked 100 images as training target images and used the remaining 50 as testing targets. We created unbiased atlases~\cite{joshi2004} for 2D and 3D from all the training data using \texttt{PyCA}\footnote{\url{https://bitbucket.org/scicompanat/pyca}} and use these atlases as our moving image(s). This allows for momentum prediction in a fixed (atlas) tangent space. We used LDDMM-shooting combined with a sum-of-squared intensity difference in \texttt{PyCA} to register the atlases to all 150 images. We chose the regularization kernel for LDDMM as $K = L^{-1} = (a\Delta^2+b\Delta+c)^{-1}$, and set $[a,b,c]$ as $[0.05, 0.05, 0.005]([1.5, 1.5, 0.15])$ for 2D(3D) images. The obtained initial momenta for the training data were used to train our network using \texttt{Torch}, 
the ones for the testing data were used for validation. We optimized the network using \textsl{rmsprop}, setting the learning rate to 0.0005, the momentum decay to 0.1 and the number of epochs to 10. We fixed the patch size to $15\times15$ for 2D and to $15\times15\times15$ for 3D. For training, we used a 1 pixel stride for the sliding window in 2D and a 7 voxel stride in 3D to keep the number of training patches manageable. For the probabilistic network, we sampled 50 times for each test case to calculate the prediction result and the variation of the deformation fields.
\begin{figure}[t!]
\begin{center}
\begin{subfigure}{1\linewidth}
\centering
\includegraphics[width=0.95\textwidth]{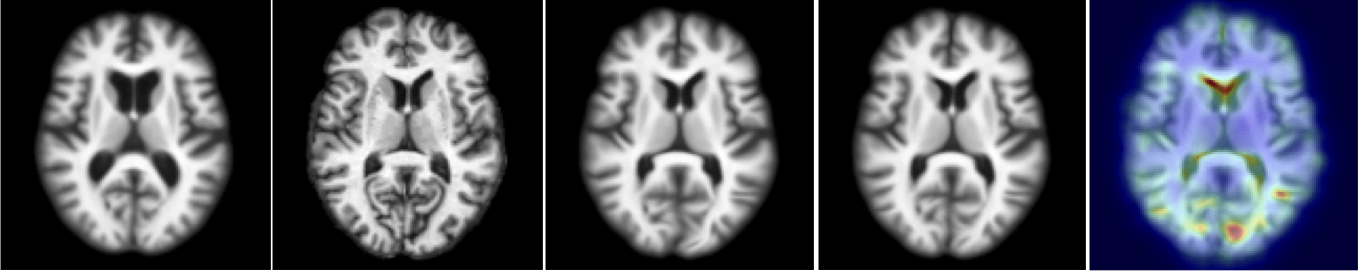}
\end{subfigure}
\vskip1ex
\begin{subfigure}{1\linewidth}
\centering
\includegraphics[width=0.95\textwidth]{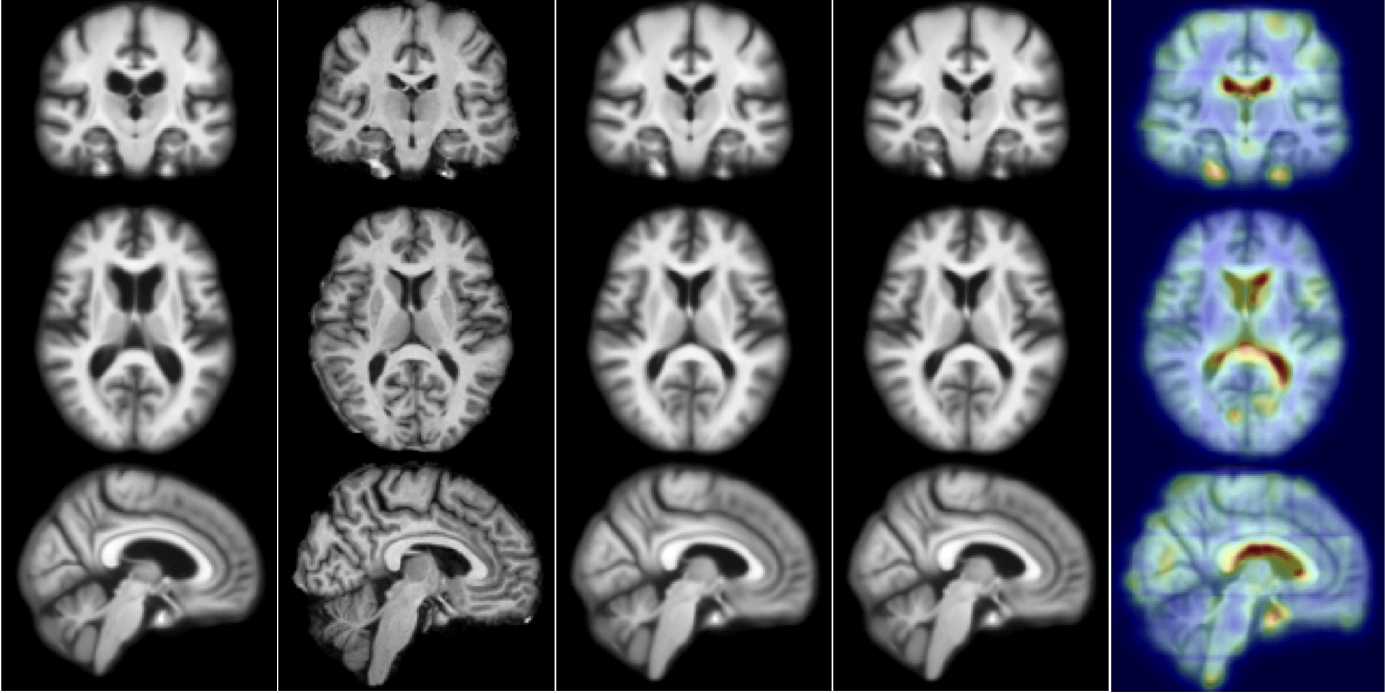}
\end{subfigure}
\vskip2ex
\caption{\label{fig:2D3D} Test example for 2D (\emph{top}) and 3D (\emph{bottom}). \emph{From left to right}: moving (atlas) image, target image, deformation result by optimizing LDDMM energy, deformation result using 50 samples from probabilistic network with a stride of 14 and patch pruning, and uncertainty as square root of the sum of the variances of deformation in all directions mapped on the predicted registration result. The colors indicates the amount of uncertainty (\textcolor{red}{red} = high uncertainty, \textcolor{blue}{blue} = low uncertainty). Best viewed in color.}
\end{center}
\vspace{-0.2cm}
\end{figure}
\vskip0.1ex
\noindent\textbf{2D Data.} For the 2D experiment, we compare our method with semi-coupled dictionary learning (SCDL), which was used to predict initial momenta for LDDMM in~\cite{Tian2015}. To compare the deformation prediction accuracy using different parameterizations, we trained (i) networks predicting the \emph{initial velocity}, $v_0=Km_0$, and the \emph{displacement field}, $\Phi(1)-\id$, of LDDMM, respectively. For the initial momentum and the initial velocity parameterizations, the resulting deformation map $\Phi(1)$ was computed by integrating Eqs.~\eqref{eqn:forward}. We quantify the deformation errors as the pixel-wise 2-norm of the deformation error with respect to the ground-truth deformation obtained by \texttt{PyCA} LDDMM. Table~\ref{table:2D3D} shows the error percentiles over all pixel and test cases. 
We observe that our initial momentum networks significantly outperform SCDL and also improve prediction accuracy compared to the initial velocity and the displacement parameterizations in both the 1-stride and the 14-stride $+$ patch pruning cases. In contrast to the initial velocity and the displacement parameterizations, both our deterministic and our probabilistic networks show comparatively small sensitivity to patch pruning and stride, validating our hypothesis that the momentum-based LDDMM parameterization is well-suited for fast predictive image registration. One of the hallmarks of LDDMM registration is that given a sufficiently strong regularization, the obtained deformation maps, $\Phi(1)$, will be diffeomorphic. To assess this property, we computed the local Jacobians of the deformation maps. Assuming no coordinate system flips, 
a diffeomorphic $\Phi(1)$ should have a positive definite Jacobian everywhere, otherwise undesirable foldings exist. 
Column `\bm{$\text{\textbf{det}} J > 0$}' of Table \ref{table:2D3D} lists the percentage of test cases with positive definite Jacobian, revealing
that our initial-momentum based networks retain this property in all scenarios, even for very large strides and patch pruning.
Direct displacement prediction, however, cannot even guarantee diffeomorphic transformations for a stride of 1 (which includes a lot of local averaging) for all our test cases and results in no diffeomorphic transformations at a stride of 14. Velocity prediction performs slightly better, but can also not guarantee diffeomorphic maps at large strides. Similarly to existing optical flow prediction methods~\cite{deepflow,flownet}, a direct prediction of displacements or velocities cannot  encode smoothness assumptions or enforce transformation guarantees. 
Our momentum parameterization encodes these assumptions by design.
Fig.~\ref{fig:2D3D} shows an example of our 2D deformation prediction with uncertainty. The predicted deformation is close to the one generated by costly LDDMM optimization. The uncertainty map shows high uncertainty at the anterior edge of the ventricle and the posterior brain cortex where drastic shape changes occur, which can be seen in the moving and the target image.

\noindent
\textbf{3D Data.} Similar to the 2D case, we computed the deformation error for every voxel in all test cases; results are listed in Table \ref{table:2D3D}. Our networks achieve sub-voxel accuracy for about 99.8\% of all the voxels. Fig.~\ref{fig:2D3D} shows one 3D registration result using the predicted deformation from our probabilistic 3D network using the mean of 50 initial momentum samples, as well as the uncertainty of the deformation field. Our prediction is able to handle large deformations. As in 2D, the uncertainty map highlights areas with drastic and ambiguous deformations.

\noindent
\textbf{Computation Speed.} On an Nvidia Titan X GPU, it took 9 hours to train a 2D network, and 72 hours to train a 3D network. By using a 14 pixel stride sliding window + patch pruning, our network (without repeated sampling) predicts the initial momentum for a 2D image in $0.19$s, and in $7.68$s for a 3D image. Compared to the GPU-based optimization in \texttt{PyCA}, we achieve an approximate speedup of $1500\mathsf{x}$/$66\mathsf{x}$ for a 2D/3D image. At a stride of 1, computational cost increases about 200-fold in 2D and 3000-fold in 3D, resulting in runtimes of about half a minute/six hours in 2D/3D. Hence, our momentum representation which is amenable to large strides is essential for achieving fast registration prediction at high accuracy while guaranteeing diffeomorphic transformations.

\vskip0.5ex
\noindent
\textbf{Discussion.} Our model is general and directly applicable to many other registration approaches with pixel/voxel-wise registration parameters (e.g., demons or elastic registration). For parametric methods (with less registration parameters) and {\it local control}, such as B-splines, we could replace the decoders by fully-connected layers. Of course, for methods where parameter locality is not guaranteed, using large stride and patch pruning may no longer be suitable. 
Future studies should assess registration accuracy in terms of landmarks and/or segmentation overlaps, compared with optimization-based techniques. Exciting extensions are: fast LDDMM-based multi-atlas segmentation; multi-modal image registration (where the input patches are from different modalities); direct image-to-image registration; fast user-interactive registration refinements (requiring prediction for a few localized patches only); and multi-patch-scale networks for better prediction. Furthermore, ambiguous deformations, caused by large deformations or appearance changes, are highlighted by the uncertainty measure, which could detect pathologies in a network trained on normals. 


\bibliographystyle{splncs03}
\vspace{-0.3cm}
\bibliography{DLMIA2016}
\end{document}